# Rule Extraction using Artificial Neural Networks

*S. M. Kamruzzaman*[1]
*Ahmed Ryadh Hasan*[2]

**Abstract**

Artificial neural networks have been successfully applied to a variety of business application problems involving classification and regression. Although backpropagation neural networks generally predict better than decision trees do for pattern classification problems, they are often regarded as black boxes, i.e., their predictions are not as interpretable as those of decision trees. In many applications, it is desirable to extract knowledge from trained neural networks so that the users can gain a better understanding of the solution. This paper presents an efficient algorithm to extract rules from artificial neural networks. We use two-phase training algorithm for backpropagation learning. In the first phase, the number of hidden nodes of the network is determined automatically in a constructive fashion by adding nodes one after another based on the performance of the network on training data. In the second phase, the number of relevant input units of the network is determined using pruning algorithm. The pruning process attempts to eliminate as many connections as possible from the network. Relevant and irrelevant attributes of the data are distinguished during the training process. Those that are relevant will be kept and others will be automatically discarded. From the simplified networks having small number of connections and nodes we may easily able to extract symbolic rules using the proposed algorithm. Extensive experimental results on several benchmarks problems in neural networks demonstrate the effectiveness of the proposed approach with good generalization ability.

**Keywords:** backpropagation, simplified network, constructive algorithm, pruning algorithm, clustering algorithm, rule extraction.

## 1. Introduction

The last two decades have seen a growing number of researchers and practitioners applying neural networks (NNs) for classification in a variety of real world applications. In some of these applications, it may be desirable to have a set of rules that explains the classification process of a trained network. The classification concept represented as rules is certainly more comprehensible to a human user than a collection of NN weights.

While the predictive accuracy obtained by NNs is often higher than that of other methods or human experts, it is generally difficult to understand how the network arrives at a particular conclusion due to the complexity of the network architecture. It is often said that a NN is practically a "black box". Even for a network with only a single hidden layer, it is generally impossible to explain why a certain pattern is classified as a member of one class and another pattern as a member of another class, due to the complexity of the network.

This paper proposes a new rule extraction algorithm, called rule extraction algorithm from artificial neural networks (REANN). A standard three-layer feedforward ANN is the basis of the algorithm. It determines automatically the number of nodes in the single hidden layer. The algorithms start with a small network (usually a single hidden neuron) and dynamically grow the network by adding and training neurons as needed until a satisfactory solution is found.

---

[1] Assistant Professor
Department of Computer Science and Engineering
Manarat International University, Dhaka-1212, Bangladesh.
Email: smk_iiuc@yahoo.com

[2] Junior Lecturer
School of Communication
Independent University Bangladesh, Chittagong, Bangladesh.
Email: ryadh78@yahoo.com

To reduce the number of rules, redundant hidden units and irrelevant input attributes are first removed by a pruning method called weight elimination algorithm and input and hidden node pruning algorithm before REANN is applied. The continuous activation function (hyperbolic tangent function) of the hidden unit is then discretize by using an efficient clustering algorithm. Rules are extracted by examining the discretized activation values of the hidden units by using a new rule extraction algorithm (REx).

## 2. Previous Work

There is quite a lot of literature on algorithms that extracts rules from trained NNs [1] [2]. Several approaches have been developed for extracting rules from a trained NN. Saito and Nakano [3] proposed a medical diagnosis expert system based on a multiplayer NN. They treated the network as black box and used it only to observe the effects on the network output caused by change the inputs.

Two methods for extracting rules from NN are described in Towell and Shavlik [4]. The first method is the subset algorithm [5], which searches for subsets of connections to a unit whose summed weight exceeds the bias of that unit. The second method, the MofN algorithm [6], clusters the weights of a trained network into equivalence classes.

H. Liu and S. T. Tan [7] proposes X2R, a simple and fast algorithm that can applied to both numeric and discrete data, and generate rules from datasets like Season-Classification, Golf-Playing that contain continuous and/or discrete data.

R. Setiono and Huan Liu [8] present a novel way to understand a neural network. Understanding a NN is achieved by extracting rules with a three phase algorithm: first, a weight decay backpropagation network is built so that important connections are reflected by their bigger weights; second, the network is pruned such that insignificant connections are deleted while its predictive accuracy is still maintained; and last, rules are extracted by recursively discretizing the hidden unit activation values.

Thrun [9] describes a rule extraction algorithm, which analyzes the input-output behavior of a network using Validity Interval Analysis. VI-Analysis divides the activation range of each network's unit into intervals, such that all network's activation values must lie within the intervals. The boundary of these intervals are obtained by solving linear programs. Two approaches of generating the rule conjectures, specific-to-general and general-to-specific, are described. The validity of these conjectures are checked with VI-analysis.

R. Setiono [10] proposes a rule extraction algorithm for extracting rules from pruned neural networks for breast cancer diagnosis. The author describes how the activation values of a hidden unit can be clustered such that only a finite and usually small number of discrete values need to be considered while at the same time maintaining the network accuracy. A small number of different discrete activation values and a small number of connections from the inputs to the hidden units will yield a set of compact rules for problem.

R. Setiono proposes a rule extraction algorithm named NeuroRule [11]. This algorithm extracts symbolic classification rule from a pruned network with a single hidden layer in two steps. First, rules that explain the network outputs are generated in terms of the discretized activation values of the hidden units. Second, rules that explain the discretized hidden unit activation values are generated in terms of the network inputs. When two sets of rules are merged, a DNF representation of network classification is obtained. Under DNF representation, the classification concept is expressed as the disjunction of or more subconcepts.

Ismail Taha and Joydeep Ghosh [12] proposes three rule extraction techniques for knowledge Based Neural Network (KBNN) hybrid systems and presents their implementation results. The suitability of each technique depends on the network type, input nature, complexity, the application nature, and the requirement transparency level. The first proposed approach (BIO-RE) is categorized as Black-box Rule Extraction (BRE) technique, while the second (Partial-RE) and third techniques (Full-RE) belong to Link Rule Extraction (LRE) category. Binarized Input-Output Rule Extraction (BIO-RE) technique extracts a set of binary rules from any NN regardless of its kind. Partial-RE extracts partial rules of most important embedded knowledge in MLP. The idea underlying Partial-RE algorithm is that it first sorts both positive and negative incoming links to all hidden and output nodes in descending order into two different sets based on their weight values. Starting from the highest positive weight (say i), it searches for individual incoming links that can cause a node j (hidden/output) to be active regardless of the other input links to this node. If such link exists, it generates a rule: If , where cf represents the measure of belief in the extracted rule and is equal to the activation value of node j. Like the Partial-RE approach, Full-RE falls in the LRE category. It is notable because: (i) It extracts rules with certainty factors from trained feedforward ANNs. (ii) It extracts all possible rules that present in the semantic interpretation of the internal structure of the trained neural network that they were extracted from. (iii) it is universal since there is no restriction on the values that any input feature can take. (iv) it is applicable to any neural network node (unit) with a monotonically increasing activation function.

R. Setiono and W. K. Leow [13] proposes a method, Fast Extraction of Rules from Neural Networks (FERNN), for extracting symbolic rules from trained feedforward neural networks with a single hidden layer. The method does not require network pruning and hence no network retraining is necessary.

R. Setiono [14] presents MofN3, a new method for extracting M-of-N rules from neural networks. The topology of the NNs is the standard three-layered feedforward networks. Units in the input layer are connected only to the units in the hidden layer, while units in the hidden layer are also connected to units in the output layer. Given a hidden unit of a trained NN with N incoming connections, show how the value of M can be easily computed. In order to facilitate the process of extracting M-of-N rules, the attributes of the dataset have binary values –1 or 1.

R. Setiono, W. K. Leow and Jack M. Zurada [15] describes a method called rule extraction from function approximating neural networks (REFANN) for extracting rules from trained neural networks for nonlinear regression. It is shown that REFANN produces rules that are almost as accurate as the original networks from which the rules are extracted. For some problems, There are sufficiently few rules that useful knowledge about the problem domain can be gained. REFANN works on a network with a single layer and one linear output unit.

## 3. Background Study
### 3.1 Introduction to Neural Networks

Artificial Neural Networks (ANNs) are relatively crude electronic models based on the neural structure of the brain. The brain basically learns from experience. It is natural proof that some problems that are beyond the scope of current computers are indeed solvable by small energy efficient packages. This brain modeling also promises a less technical way to develop machine solutions. This new approach to computing also provides a more graceful degradation during system overload than its more traditional counterparts.

A neural network is a massively parallel-distributed processor made up of simple processing units, which has a natural propensity (tendency) for storing experimental knowledge and making it available for use.
Neural network are also referred to in literature as neurocomputers, connectionist networks, parallel-distributed processors, etc [16].

A typical neural network is shown in the following figure where input, hidden and output layers are arranged in a feedforward manner.

Input Layer   Hidden Layer   Output Layer

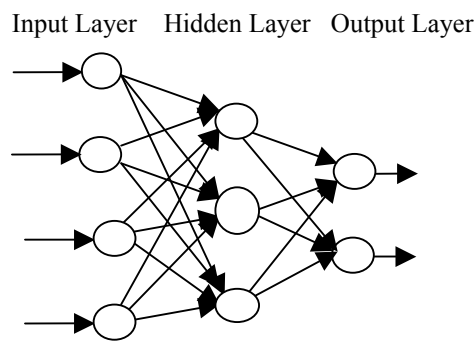

Fig 3.1: A simple neural network with input nodes in the input layer, hidden nodes in the hidden layer and output node in the output layer.

### 3.2 Constructive Algorithm

Constructive (or generative) algorithms offer an attractive framework for the incremental construction of near-minimal neural-network architectures. These algorithms starts with a small network (usually a single neuron) and dynamically grow the network by adding and training neurons as needed until a satisfactory solution is found [17] [18].
The algorithm starts with one unit in the hidden layer. Additional units are added to the hidden layer one at a time to improve the accuracy of the network on the training data.

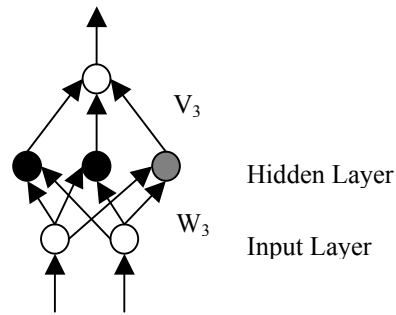

Fig: 3.2 Feedforward neural network with 3 hidden units.

When a third hidden unit is added, the optimal weights obtained from the original network with two hidden units are used as initial weights for retraining the new network. The initial value for w3 and v3 are chosen randomly. Artificial neural networks have been successfully applied to solve problems in pattern classification, function approximation, optimization, pattern matching and associative memories. Multilayer Feedforward networks trained using the Backpropagation learning algorithm is limited to search for a suitable set of weights in an a priori fixed network topology. The traditional Backpropagation method is the need to determine the number of units in the hidden layer prior to training. To overcome this difficulty, many algorithms that construct a network dynamically have been proposed.

Network pruning offers another approach for dynamically determining an appropriate network topology. Pruning techniques begin by training a larger than necessary network and then eliminate weights and neurons that are deemed redundant. Constructive algorithms offer several significant advantages over pruning-based algorithms including, the ease of specification of the initial network topology, better economy in terms of training time and number of training examples, and potential for converging to a smaller network with superior generalization.

### 3.3 Pruning a Feedforward Neural Network

Network pruning offers another approach for dynamically determining an appropriate network topology. Pruning techniques [19] begin by training a larger than necessary network and then eliminate weights and neurons that are deemed redundant. Typically, methods for removing weights involve adding a penalty term to the error function [20]. It is hoped that adding a penalty term to the error function, unnecessary connection will have smaller weights and therefore complexity of the network can be significantly reduced. This paper aims at pruning the network size both in number of neurons and number of interconnections between the neurons. The pruning strategies along with the penalty function are described in the subsequent sections.

### 3.4 Penalty Function

When a network is to be pruned, it is a common practice to add a penalty term to the error function during training [20]. Usually, the penalty term, as suggested in different literature, is

$$P(w,v) = \varepsilon_1 \left( \sum_{m=1}^{h} \sum_{l=1}^{n} \frac{\beta(w_l^m)^2}{1+\beta(w_l^m)^2} + \sum_{m=1}^{h} \sum_{p=1}^{o} \frac{\beta(v_p^m)^2}{1+\beta(v_p^m)^2} \right) + \varepsilon_2 \left( \sum_{m=1}^{h} \sum_{l=1}^{n} (w_l^m)^2 + \sum_{m=1}^{h} \sum_{p=1}^{o} (v_p^m)^2 \right) \quad (1)$$

Given an n-dimensional example $x^i, i\varepsilon\{1,2,....,k\}$ as input, let $w_l^m$ be the weight for the connection from input unit $l, l\varepsilon\{1,2,....,n\}$ to hidden unit $m, m\varepsilon\{1,2,....,h\}$ and $v_p^m$ be the weight for the connection from hidden unit m to output unit $p, p\varepsilon\{1,2,....,o\}$, the $p^{th}$ output of the network for example $x^i$ is obtained by computing

$$S_p^i = \sigma\left(\sum_{m=1}^{h} \alpha^m v_p^m\right), \text{ where} \quad (2)$$

$$\alpha^m = \delta\left(\sum_{l=1}^{n} x_l^i w_l^m\right), \delta(x) = (e^x - e^{-x})/(e^x - e^{-x}) \quad (3)$$

The target output from an example $x^i$ that belongs to class $C_j$ is an o-dimensional vector $t^i$, where $t_p^i = 0$ if $p = j$ and $t_p^i = 1, j, p = 1,2…o$. The back propagation algorithm is applied to update the weights (w, v) and minimize the following function:

$$\theta(w,v) = F(w,v) + P(w,v) \quad (4)$$

where $F(w,v)$ is the cross entropy function as defined

$$F(w,v) = -\sum_{i=1}^{k} \sum_{p=1}^{o} \left(t_p^i \log S_p^i + (1-t_p^i)\log(1-S_p^i)\right) \quad (5)$$

and $P(w,v)$ is a penalty term as described in (1) used for weight decay.

## 4. Proposed Algorithm

### 4.1 Overview of the Approach

This paper proposes a new rule extraction algorithm, called rule extraction algorithm from artificial neural networks (REANN). A standard three-layer feedforward ANN is the basis of the algorithm. It determines automatically the number of nodes in the single hidden layer. The algorithms start with a small network (usually a single hidden neuron) and dynamically grow the network by adding and training neurons as needed until a satisfactory solution is found.

To reduce the number of rules, redundant hidden units and irrelevant input attributes are first removed by a pruning method called weight elimination algorithm and input and hidden node pruning algorithm before REANN is applied. The continuous activation function (hyperbolic tangent function) of the hidden unit is then discretize by using an efficient clustering algorithm. Rules are extracted by examining the discretized activation values of the hidden units by using a new rule extraction algorithm (REx).

### 4.2 The REANN Algorithm

A standard three-layer feedforward ANN is the basis of the algorithm. A basic constructive algorithm is used in order to determine the appropriate number of hidden units. Weight decay is implemented while backpropagation is carried out. A basic pruning algorithm is used for detecting the relevant inputs and connections. The hyperbolic tangent function $f(x) = \dfrac{e^x - e^{-x}}{e^x + e^{-x}}$ is used as the hidden unit activation function; the hidden unit activation of a neural network can take any value in the interval [-1, 1]. An efficient method is used for discretizing the output of hidden nodes. Rules are extracted by examining the discretized activation values of the hidden unit.

The algorithm first discretizes the activation values of hidden nodes into a manageable number of discrete values without sacrificing the classification accuracy of the network. A small set of the discrete activation values make it possible to determine both the dependency among the output values and the hidden node values and the dependency among hidden node activation values and input values. From the dependencies rules can be generated.

The aim of this work is to search for simple rules with high predictive accuracy. The basic idea of the proposed algorithm is this: using first order information in the data to determine shortest sufficient conditions in a pattern (i.e. the rule under consideration) that can differentiate the pattern from patterns of other classes and prune redundant rules. The sole use of first order information avoids the combinatorial complexity in computation, although it is well that using higher order information may provide better results.

**4.3 Detailed Algorithm**

1. Create an initial ANN with a single hidden layer. Initially hidden layer contains only one node. Randomly initialize connection weights within a certain range.
2. Train the network using BP.
3. Determine the appropriate number of hidden nodes by using a basic constructive algorithm.
4. To detect the relevant inputs, remove redundant connections in the network using a basic pruning algorithm.
5. Use an efficient method and a cluster algorithm for discretizing the output of hidden nodes.
6. Using the discretize values as input, apply REx to generate rules that describe network outputs.
7. For each hidden units, extract rules using REx to describe its activation values in terms of the inputs.
8. Generate rules that relate the inputs and the outputs by combining rules generated in last two Steps.
9. Prune redundant rules.

**4.4 Rule Extraction Algorithm (REx)**

The pseudo code of the rule extraction algorithm (REx) is given below. The REx is composed of three major functions:
1. Rule Extraction: this function iteratively generates shortest rules and remove/marks the patterns covered by each rule until all patterns are covered by the rules.
2. Rule Clustering: rules are clustered in terms of their class levels and
3. Rule Pruning: redundant or more specific rules in each cluster are removed.

A default rule should be chosen to accommodate possible unclassifiable patterns. If rules are clustered, the choice of the default rule is based on clusters of rules.

Rule Extraction algorithm (REx) is given below:

**Rule Extraction:**
    i=0; while (data is NOT empty/marked){
    generate Ri to cover the current pattern and differentiate it from patterns in other categories;
    remove/mark all patterns covered by Ri ; i++}

**Rule Clustering:**
    cluster rules according to their class levels;

**Rule Pruning:**
    replace specific rules with more general ones;
    remove noise rules;
    eliminate redundant rules;
determine a default rule.

**4.5 Flowchart of REANN**

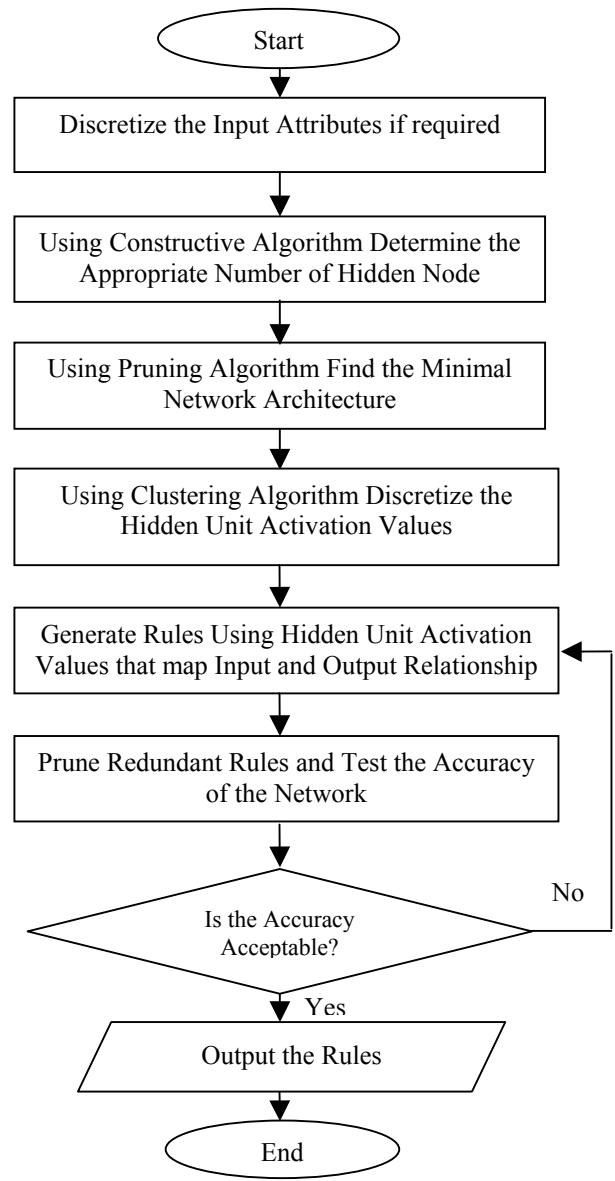

**Fig.1:** Flow chart of the REANN algorithm

## 5. Experimental Studies

We have selected the most well known benchmarks to test the REANN described in the previous section. These benchmarks are the Cancer, Season Classification, Golf-Playing problems. The data sets representing all these problems were real world data and obtained from the UCI machine learning benchmark repository [21]. The characteristics of the data sets are summarized in Table I. The detailed descriptions of the data sets are available at ics.uci.edu (128.195.11) in directory/pub/machine-learning-databases. Figure 2 shows the training error for the breast cancer data set and Figure 3 shows the hidden unit addition for diabetes data set.

### 5.1 Classification Problems
#### i) The Breast Cancer Problem
Diagnosis of breast cancer. Try to classify a tumor as either benign or malignant based on cell descriptions gathered by microscopic examination. Input attributes are for instance the clump thickness, the uniformity of cell size and cell shape, the amount of marginal adhesion, and the frequency of bare nuclei.

The data set representing this problem contained 699 examples. Each example consisted of nine-element real valued vectors. This was a two-class problem. All inputs are continuous; 65.5% of the examples are benign. This makes for an entropy of 0.93 bits per example. This dataset was created based on the breast cancer Wisconsin" problem dataset from the UCI repository of machine learning databases.

#### ii) Season Classification Problem
The season classification data set contains discrete data only. There were 11 examples in the data set, each of which consisted of three-elements. These are weather, tree and temperature. This was a four-class problem.
#### iii) Golf Playing Problem
The Golf playing data set contains both numeric and discrete data only. There were 14 examples in the data set, each of which consisted of four-elements. These are outlook, temperature, humidity and wind. This was a two-class problem. Table 1 shows the characteristics of the data sets.

**Table 1:** Characteristics of data sets

| Data Sets | No. of Examples | Attributes | Classes |
|---|---|---|---|
| Breast Cancer | 699 | 9 | 2 |
| Diabetes | 768 | 8 | 2 |
| Iris | 150 | 4 | 3 |
| Wine | 178 | 13 | 3 |
| Season Classification | 11 | 3 | 4 |
| Golf Playing | 14 | 4 | 2 |
| Lenses | 24 | 4 | 3 |

### 5.2 Experimental Setup

In all experiments, one bias unit with a fixed input 1 was used for hidden and output layers. The learning rate was set between [0.1, 1.0] and the weights were initialized to random values between [-1.0, 1.0]. Value of ε for clustering was set between [0.1, 1.0]. Values of weight decay parameters $\varepsilon_1$, $\varepsilon_2$, were set between [0.05, .5] and [$10^{-4}$, $10^{-8}$] and β was 10 for penalty function. Hyperbolic tangent function $f(x) = \frac{e^x - e^{-x}}{e^x + e^{-x}}$ is used as hidden unit activation function and logistic sigmoid function $f(x) = \frac{1}{1 + e^{-x}}$ as output unit activation function.

In this study, all data sets representing the problems are divided into two sets. One is the training set and the other is the test set. Note that no validation set is used in this study. The numbers of examples in the training set and test set are based on numbers in other works, in order to make comparison with those works possible. First 50% of the total data was used for training and rest 50% was used as test data.

### 5.3 Pruned Networks

**Breast cancer diagnosis problem:** The smallest of the pruned network for breast cancer diagnosis problem with more than 93% accuracy rate on the training data has only 1 hidden unit and 5 connections. The network is depicted in Figure 1. The accuracy on the training data and testing data are 93.429% and 96.275% respectively. In this example only $I_1$, $I_6$ and $I_9$ are important. Only three discrete values are needed to maintain the accuracy of the network. The values found by the clustering algorithm are 0.987, -0.986 and 0.004. Of the 350 training data, 238 patters have the first value, 106 have the second value and rest 6 patterns have third value. The connection from the hidden unit to the first output unit is 3.0354 and to the second output unit is −3.0354.

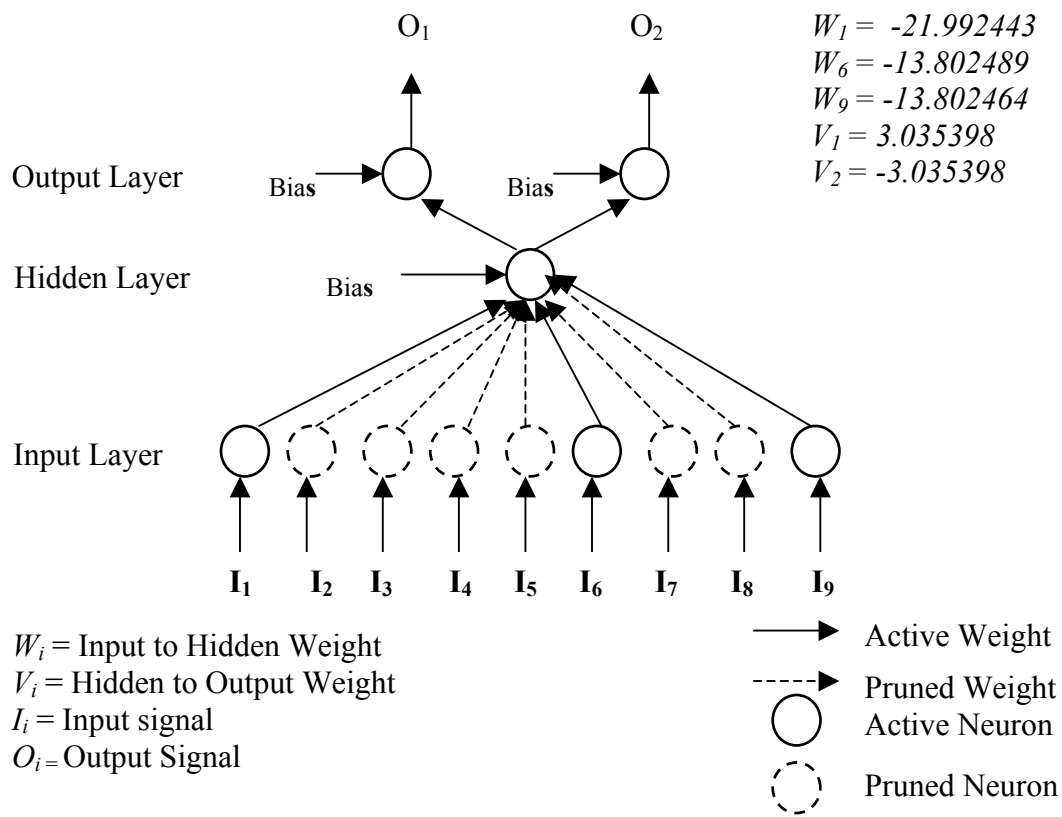

**Fig. 1:** A pruned network for breast cancer diagnosis problem. The accuracy on training and test data sets are 93.429% and 96.275% respectively.

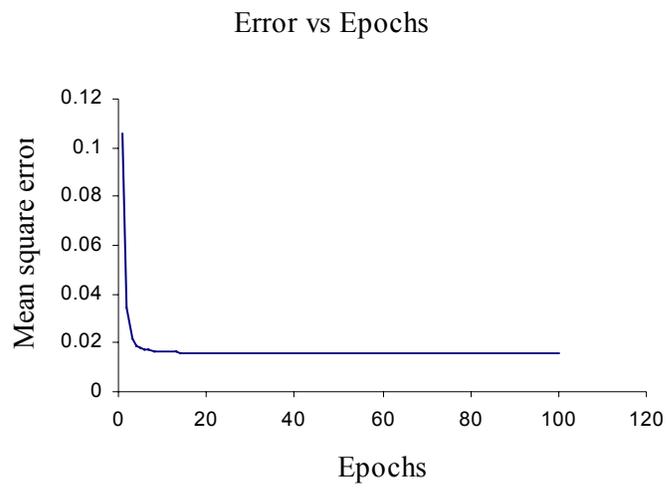

**Fig. 2:** Training time error of a network (9-1-1) for Cancer data set.

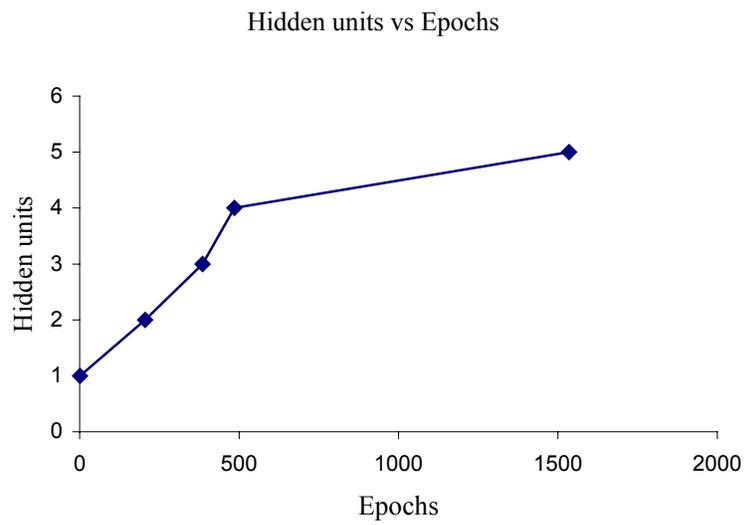

**Fig 3:** Hidden unit addition for the diabetes data set.

**5.4 Extracted Rules**

**For Breast Cancer Problem:**

**Rule 1:** If Clump thickness (A1) <= 0.6 and Bare nuclei (A6) <= 0.5
and Mitosis(A9) <= 0.3, then benign
**Default Rule:** malignant.

**For Season Classification Data Set:**

**Rule 1:** If Tree (A2) = yellow then autumn
**Rule 2:** If Tree (A2) = leafless then autumn
**Rule 3:** If Temperature (A3) = low then winter
**Rule 3:** If Temperature (A3) = high then summer
**Default Rule:** spring

**For Golf Playing Problem:**
**Rule 1:** If Outlook (A1) = sunny and Humidity>=85 then don't play
**Rule 2:** Outlook (A1) = rainy and Wind= strong then don't play
**Default Rule:** play

**5.5 Rules Accuracy**

Table 2 shows the rule accuracy of the extracted rules for several data sets.

**Table 2:** Rule accuracy

| Data Set | Training Set | Test Set |
|---|---|---|
| Breast Cancer | 327/350 93.43 % | 336/349 96.28 % |
| Season Classification | 11/11 100 % | 11/11 100 % |
| Golf Playing | 14/14 100 % | 14/14 100 % |

**6. Discussion and Comparison with Related works**

We have shown how rules can be extracted from a trained neural network without making any assumptions about the network's activations or having initial knowledge about the problem domain. If some knowledge is available, however, it can always be incorporated into the network. For example, connections in the network from inputs thought to be not relevant can be given large penalty parameters during training, while those thought to be relevant can be given zero or small penalty parameters. Our algorithm does not require threshold activation function to force the activation values to be zero or one, nor does it require the weights of the connections to be restricted in a certain range. Table 3 shows a comparative study of the performance of our algorithm with the works of other researchers.

**Table 2:** Comparative study

| Data Set | Feature | REANN | NN RULES | DT RULES | C4.5 | NN-C4.5 | OC1 | CART |
|---|---|---|---|---|---|---|---|---|
| **Breast Cancer** | No. of Rules | 2 | 4 | 7 | | | | |
| | Avg. No. of Conditions | 3 | 3 | 1.75 | | | | |
| | Accuracy % | **96.28** | 96 | 95.5 | 95.3 | 96.1 | 94.99 | 94.71 |
| **Season Classification** | No. of Rules | | RULES | RULES-2 | X2R | | | |
| | | 5 | 7 | | 6 | | | |
| | Avg. No. of Conditions | 1 | 2 | | 1 | | | |
| | Accuracy % | **100.0** | 100.0 | | 100.0 | | | |
| **Golf Playing** | No. of Rules | 3 | 8 | 14 | 3 | | | |
| | Avg. No. of Conditions | 2 | 2 | 2 | 2 | | | |
| | Accuracy % | **100.0** | 100.0 | 100.0 | 100.0 | | | |

## 7. Conclusions

Neural networks are often viewed as black boxes. While their predictive accuracy is high, one usually cannot understand why a particular outcome is predicted. In this paper, we have attempted to open up these black boxes. Two factors make this possible. The first is a robust pruning algorithm. Using penalty function, we have been to prune networks such that only very few input units, hidden units and connections left in the networks. By eliminating redundant weights, redundant input and hidden units are identified and removed from the network. Removal of these redundant units significantly simplifies the process of rule extraction and the extracted rules themselves. The second factor is the clustering of the hidden unit activation values. The fact that the number of distinct activation values at the hidden units can be made small enough enables us to extract simple rules.

We describe REANN, an algorithm that can extract rules from a standard feedforward neural network. Network training and pruning is done via the simple and widely used backpropagation method. No restriction is imposed on the activation values of hidden units or output units. An important feature of our rule extraction algorithm is its recursive nature. Extract rules are a one-to-one mapping of the network. They are compact and comprehensible, and do not involve any weight values. The accuracy of the rules from a pruned network is as high as the accuracy of the network.

Extensive experimental results on several benchmarks problems in neural networks including Cancer, Season Classification, Golf-Playing demonstrate the effectiveness of the proposed approach with good generalization ability.

**References**

[1] R. Andrews, J. Diederich and A. B., Tickle, "Survey and critique of techniques for extracting rules from trained artificial neural networks," Knowledge Based System, vol. 8, no. 6, pp. 373-389, 1995.

[2] Ashish Darbari, "Rule extraction from ANN: a survey," July 2001.

[3] K. Saito and R. Nakano, "Medical diagnosis expert system based on PDP model," Proceedings of IEEE International Conference on Neutal Networks, IEEE Press, New York, pp. 1255-1262, 1988.

[4] G. G., Towell and J. W., Shavlik, "Extracting refined rules from knowledge-based system neural networks," Machine Learning, vol. 13, no. 1, pp. 71-101, 1993.

[5] L. Fu, "Rule learning by searching on adapted nets, " Proceedings of the Ninth National Conference on Artificial Intelligence, AAAI Press/ The MIT Press, Menlo Park, CA, pp. 590-595, 1991.

[6] G. G., Towell and J. W., Shavlik, "Knowledge-based artificial neural networks," Artificial Intelligence, vol. 70, pp. No. 1, 2, pp. 119-165, 1994.

[7] H. Liu and S. T. Tan, "X2R: A fast rule generator," Proceedings of IEEE International Conference on Systems, Man and Cybernetics, Vancouver, CA, 1995.

[8] Rudy Setiono and Huan Liu, "Understanding neural networks via rule extraction," Proceedings of the 14$^{th}$ International Joint Conference on Artificial Intelligence, pp. 480-485, 1995.

[9] S. Thrun, "Extracting rules from artificial neural networks with distributed presentations, " Advances in Neural Information Processing Systems, vol. 7, The MIT Press, Cambridge, MA, 1995.

[10] R. Setiono, "Extracting rules from pruned neural networks for breast cancer diagnosis," Artificial Intelligence in Medicine, vol. 8, no. 1, pp. 37-51, February 1996.

[11] R. Setiono and H. Liu, "Symbolic presentation of neural networks," IEEE Computer, pp. 71-77, March 1996.

[12] Taha and J. Ghosh, "Three Techniques for extracting rules from feedforward networks", 1996.

[13] R. Setiono and W. K. Leow, " FERNN: An algorithm for fast extraction of rules from neural networks," Applied Intelligence, vol. 12, no. ½, pp. 15-25, 2000.

[14] R. Setiono, "Extracting M-of-N rules from trained neural networks," IEEE Transactions of Neural Networks, vol. 11, no. 2, pp. 512-519, 2000.

[15] R. Setiono, W. K. Leow and Jack M. Zurada, "Extraction of Rules from Artificial Neural Networks for Nonlinear regression," IEEE Transactions of Neural Networks, vol. 13, no. 3, pp. 564-577, May 2002.

[16] Simon Haykin, "Neural Networks: A comprehensive foundation," Pearson Education Asia, Third Indian Reprint 2002.

[17] R. Setiono and L.C.K. Hui, "Use of quasi-Newton method in a feedforward neural network construction algorithm", IEEE Trans. Neural Networks, vol. 6, no.1, pp. 273-277, Jan. 1995.

[18] R. Parekh, J.Yang, and V. Honavar, "Constructive Neural Network Learning Algorithms for Pattern Classification", IEEE Trans. Neural Networks, vol. 11, no. 2, March 2000.

[19] R. Setiono, Huan Liu, "Improving Backpropagation Learning with Feature Selection", Applied Intelligence, vol. 6, no. 2, pp. 129-140, 1996.

[20] R. Setiono, Huan Liu, "Understanding Neural networks via Rule Extraction", In Proceedings of the International Joint conference on Artificial Intelligence, pp. 480-485, 1995.

[21] L. Prechelt, "Proben1-A Set of Neural Network Benchmark Problems and Benchmarking Rules", University of Karlsruhe, Germany, 1994.